\documentclass[11pt,a4paper]{article}
\usepackage{times,latexsym}
\usepackage{url}
\usepackage[T1]{fontenc}

\usepackage[table]{xcolor} 
\usepackage[acceptedwithA]{tacl2021v1}

\usepackage{multirow}
\usepackage{booktabs} 
\usepackage{array} 
\usepackage{amsmath} 
\usepackage{amssymb} 
\usepackage{graphicx} 
\usepackage{colortbl} 
\usepackage{makecell} 
\usepackage{threeparttable}

\def\equationautorefname~#1\null{ 
  Equation~(#1)\null
}

\title{Fine-tuning Large Language Models with Limited Data: A Survey and Practical Guide}

\author{
  Marton Szep$^{1,2}$
  \quad
  Daniel Rueckert$^{1,3,4}$
  \quad
  Rüdiger von Eisenhart-Rothe$^2$
  \quad
  Florian Hinterwimmer$^{1,2}$
  \\
  \ \\
  $^1$Chair for AI in Healthcare and Medicine, Technical University of Munich (TUM) and\\TUM University Hospital, Munich, Germany
  \\
  $^2$Department of Orthopaedics and Sports Orthopaedics, TUM University Hospital, Germany 
  \\
  $^3$Munich Center for Machine Learning (MCML), Munich, Germany
  \\
  $^4$Department of Computing, Imperial College London, UK
  \\
  \href{mailto:marton.szep@tum.de}{\textcolor{black}{\texttt{marton.szep@tum.de}}}
}

\date{}

\begin{document}
\maketitle
\begin{abstract}
Fine-tuning large language models (LLMs) with limited data poses a practical challenge in low-resource languages, specialized domains, and constrained deployment settings. While pre-trained LLMs provide strong foundations, effective adaptation under data scarcity requires focused and efficient fine-tuning techniques.
This paper presents a structured and practical survey of recent methods for fine-tuning LLMs in data-scarce scenarios.
We systematically review parameter-efficient fine-tuning techniques that lower training and deployment costs, domain and cross-lingual adaptation methods for both encoder and decoder models, and model specialization strategies. 
We further examine preference alignment approaches that guide model behavior using limited human or synthetic feedback, emphasizing sample and compute efficiency.
Throughout, we highlight empirical trade-offs, selection criteria, and best practices for choosing suitable techniques based on task constraints, including model scaling, data scaling, and the mitigation of catastrophic forgetting.
The aim is to equip researchers and practitioners with actionable insights for effectively fine-tuning LLMs when data and resources are limited.
\end{abstract}

\section{Introduction}
Pre-trained language models (PLMs) have driven unprecedented advances in NLP, showing strong capabilities across a wide range of tasks \citep{han_pre-trained_2021}. 
However, their development and adaptation typically require large-scale data and computational resources, often unavailable in real-world scenarios, particularly in low-resource languages or specialized domains such as medicine \citep{crema_advancing_2023}, law \citep{noguti_small_2023}, chemistry \citep{jablonka_leveraging_2024}, or finance \citep{zhao_bert_2021}.
To address these constraints, transfer learning has become the standard paradigm: models are first pre-trained on general corpora and then adapted through fine-tuning (FT) and preference alignment (PA). 
Yet, these adaptation stages can still demand significant annotated data, and naive FT under low-data conditions risks overfitting, catastrophic forgetting \citep{ramasesh_effect_2021}, and poor generalization.

Given the rapidly evolving field and broad array of methods, a structured and pragmatic review is necessary to navigate current strategies and make informed choices under real-world constraints.
This paper offers a practical guide to fine-tuning large language models under data scarcity.
We examine recent developments across four key areas. First, we review \textbf{(1)~Parameter-efficient Fine-tuning} (PEFT, \autoref{sec:param-eff}), which enables modular and cost-effective adaptation. 
Second, we explore \textbf{(2)~Domain and Cross-lingual FT} (\autoref{sec:ft-generalization}), targeting robustness to distributional shifts. 
Third, \textbf{(3)~FT for Specialization} (\autoref{sec:ft-specialization}) covers techniques for specialized tasks, domains, and low-resource languages, including scaling, data efficiency, and design recommendations. 
Fourth, we analyze \textbf{(4)~Preference Alignment} (PA \autoref{sec:pref-align}), including data requirements, scaling behavior, and downstream effects.
We conclude with a focused discussion on practical challenges and emerging strategies, such as the trade-offs between encoder and decoder architectures, cross-lingual transfer in low-resource contexts, and the potential of model merging. Our emphasis throughout is on actionable insights and contextualized, concrete guidance to support effective LLM fine-tuning when data is limited.

\section{Related Surveys}
Several prior works address related aspects or have partial overlap with our work.
Some surveys target data augmentation as a solution to data scarcity \citep{feng_survey_2021, chen_empirical_2023}, including domain-specific generation techniques for low-resource settings \citep{stylianou_domain-aligned_2023}.
Model-centric strategies are covered in surveys on PEFT methods \citep{han_parameter-efficient_2024, lialin_scaling_2024}, while \citet{treviso_efficient_2023, wan_efficient_2024, xu_resource-efficient_2025} offer broader coverage of efficient training techniques across different stages of model development.
Other works focus on low-resource supervised NLP \citep{hedderich_survey_2021}, instruction tuning \citep{zhang_instruction_2024}, and PA \citep{gao_towards_2024, jiang_survey_2024}.
In contrast, our work offers a pragmatic overview of the full post-pretraining pipeline for adapting PLMs with limited data. We cover PEFT, domain and cross-lingual generalization, specialization strategies, and PA in an integrated survey, emphasizing practical insights, comparative trade-offs, and methodological guidance to support informed decisions in data-scarce scenarios.

\section{Parameter-efficient Fine-tuning}
\label{sec:param-eff}

Fine-tuning (FT) all parameters of large PLMs is computationally expensive, sample-inefficient, and often unstable in low-resource regimes \citep{dodge_fine-tuning_2020}.
Parameter-efficient fine-tuning (PEFT) mitigates these issues by updating only a small fraction of weights, enabling strong adaptation with lower cost and reduced risk of catastrophic forgetting in data-scarce scenarios.
We organize PEFT methods by how they \emph{compose} with the base model parameters  \citep[following][]{pfeiffer_modular_2023}, and highlight key performance and efficiency considerations for selecting the appropriate method for a given task and data setup.
For comprehensive surveys, see \citet{han_parameter-efficient_2024,pfeiffer_modular_2023}.

\subsection{Parameter composition}
\label{sec:param-comp}
These methods modify existing weights through sparse or low-rank updates.

\vspace{3pt}\noindent\textbf{Selective methods} \label{par:selective} train only a subset of the model weights (specific layers, parameter types, etc.) while keeping the rest fixed.
They build on the inductive bias of sparsity: in an over-parameterized model, a small fraction of parameters plays a disproportionately important role for a particular task.
Variants include tuning specific (usually last) layers or model components like the bias, or LayerNorm parameters \citep{devlin_bert_2019,ben_zaken_bitfit_2022,liu_few-shot_2022}.
Another line of work optimizes subnetworks based on gradient information either by selecting a specific subnetwork \citep{xu_raise_2021,ansell_composable_2022,sung_training_2021}, or adaptively choosing it through a bi-level optimization strategy \citep{zhang_fine-tuning_2022,yu_unlearning_2023}. 
These methods are extremely lightweight and overhead-free at inference time, but usually underperform compared to approaches that introduce new trainable parameters.

\vspace{3pt}\noindent\textbf{Reparametrization methods} \label{par:reparametrization} replace full weight updates with low-dimensional structures, motivated by the low-rank nature of PLM weights \citep{aghajanyan_intrinsic_2021}.
The most widely used approach, LoRA \citep{hu_lora_2021}, decomposes weight updates into the product of two low-rank matrices and can be merged into the backbone at inference, avoiding extra latency.
Its effectiveness led to numerous extensions improving efficiency, robustness to hyperparameters, and calibration \citep{liu_dora_2024,dettmers_qlora_2023,hayou_lora_2024,ding_sparse_2023,zhang_adaptive_2022,edalati_krona_2022}.
Reparametrization methods work best on larger models and with moderate to high data sizes, and often match full FT performance with careful tuning \citep{van_veen_radadapt_2023}.

\subsection{Input composition}
\label{sec:input-comp}
These methods extend the input with learned embeddings that act as task-specific context.

\vspace{3pt}\noindent\textbf{Soft prompt} \label{par:soft_prompt} 
prepend learned embedding vectors to the input or key-value pairs in attention layers \citep{li_prefix-tuning_2021,lester_power_2021}. 
They are a popular choice to adapt models across domains and languages due to their flexibility and efficiency \citep{tu_efficiently_2024,goswami_switchprompt_2023,zhao_adpl_2022}, but often converge slowly and perform reliably only at very large model scales \citep{mao_unipelt_2022}.
Variants such as $\mathtt{IA^3}$ \citep{liu_few-shot_2022} improve efficiency by learning vectors to scale hidden activations, achieving adapter-level performance with fewer parameters.
Soft prompts typically incur additional inference overhead, making them more suitable for scenarios prioritizing training efficiency over runtime cost.

\subsection{Function composition}
\label{sec:func-comp}
These methods augment the frozen backbone with lightweight trainable modules. 

\vspace{3pt}\noindent\textbf{Adapters} \label{par:adapters} are small feedforward modules inserted between transformer layers \citep{houlsby_parameter-efficient_2019, pfeiffer_mad-x_2020}. 
They reduce training memory and enable flexible modular composition across tasks \citep{pfeiffer_adapterfusion_2021,wang_adamix_2022,chronopoulou_adaptersoup_2023}, but sequential insertion slows down inference by up to 40\% \citep{ruckle_adapterdrop_2021}.
Parallel and Ladder Side-tuning variants mitigate this issue \citep{he_towards_2021,sung_lst_2022}.
Adapters can act as bottlenecks, requiring wider layers and more paramaters than other PEFT methods to maintain performance \citep{hu_llm-adapters_2023}.
Their efficiency can be improved through low-rank reparametrization, linking them conceptually with methods in \autoref{par:reparametrization}.

\subsection{Hybrid methods} \label{par:hybrid} 
Hybrid methods combine different forms of composition to leverage complementary strengths. 
Examples include Compacter \citep{karimi_mahabadi_compacter_2021}, which integrates adapters with low-rank decomposition, and UniPELT, which dynamically activates adapters, prefixes, and LoRA \citep{mao_unipelt_2022}. 
Such methods can be powerful in very low-resource settings, but are generally more sensitive to hyperparameter tuning \citep{chen_parameter-efficient_2022}.

\subsection{Selection criteria} 
Choosing the right PEFT method depends on task type, model size, and data scale.
Full FT becomes favorable only with million-scale datasets \citep{zhang_when_2023}, whereas PEFT often matches or outperforms it under 100k samples \citep{lialin_scaling_2024,ding_parameter-efficient_2023,jukic_parameter-efficient_2023}. 
In practice, selective and reparametrization methods update less than 1\% of parameters, while adapters and soft prompts can require up to several percent \citep{lialin_scaling_2024}.
Selective methods are simple, efficient, but they usually underperform \citep{liu_few-shot_2022}, with LayerNorm tuning sometimes found effective \citep{lialin_scaling_2024}.
Soft prompts can outperform LoRA on the scale of a few thousand samples \citep{liu_few-shot_2022,zhang_when_2023,ding_parameter-efficient_2023}, but converge slowly and are fragile at smaller model scales. 
Adapters and LoRA stand out as the most robust choices, reliably matching full FT with minimal hyperparameter tuning in moderate- to large-scale data settings \citep[over 10k samples;][]{zhang_when_2023}.
Hybrid methods can yield gains in extremely low-resource scenarios but at the cost of added complexity \citep{chen_parameter-efficient_2022,mao_unipelt_2022,he_towards_2021}.

Inference overhead varies across methods, with adapters imposing the largest slowdown due to added depth, followed by soft prompts (30–40\%), reparametrization methods (although they can usually be merged into the model), and selective methods, which incur minimal to no runtime cost \citep{hu_lora_2021,lialin_scaling_2024}.
\autoref{tab:nlp_tasks} and \autoref{sec:task-specific} offer more practical guidance on selecting PEFT methods across discriminative tasks for encoder models.


\subsection{Efficient design choices}
Efficient design choices in PEFT methods are vital for maximizing performance gains while minimizing resource requirements.
Studies show that PEFT methods exhibit better performance and convergence rate as the model size increases, also progressively narrowing the gap between full FT and PEFT \citep{zhang_when_2023,ding_parameter-efficient_2023,lialin_scaling_2024}.
Intuitively, with increasing data size expressivity should be expanded through more PEFT parameters. 
However, studies \citep{zhang_when_2023,ding_parameter-efficient_2023} show that the gains of scaling the tunable parameters (e.g., LoRA rank $4\rightarrow128$ or prompt-tuning prompt length  $50\rightarrow600$) are marginal when scaling the data size up to 100k samples, especially compared to scaling the model. 
With a fixed compute budget, it is preferable to opt for larger models with reduced precision, e.g. 8 or 4-bit quantization \citep{dettmers_qlora_2023}.

\section{Domain and Cross-lingual Fine-tuning} \label{sec:ft-generalization}

\begin{table*}[!ht]
  \centering
  \scriptsize
  \rowcolors{2}{white}{gray!10}
  \begin{tabular}{>{\cellcolor{white}}m{3mm}>{\centering\arraybackslash}m{2.2cm}|>{\centering\arraybackslash}m{1.9cm}|>{\raggedright\arraybackslash}m{3cm}|>{\raggedright\arraybackslash}m{3.1cm}|>{\raggedright\arraybackslash}m{3.2cm}}
  \toprule
  & \textbf{Method} & \textbf{Requirements} & \centering\textbf{Advantages} & \centering\arraybackslash\textbf{Limitations} & \centering\arraybackslash\textbf{Use Cases} \\ \midrule

  & Continued Pre-training (\autoref{sec:continued-pretrain}) & Monolingual, in-domain corpora & Can use any kind of relevant unlabeled corpora & Computationally intensive; Can lead to catastrophic forgetting  & Cross-lingual or domain-specific adaptation \\ 
  & Pattern-exploiting training (\autoref{par:pattern-exploit}) & Few labeled examples & Better alignment of FT and PT task & Handcrafting; Slow autoregressive decoding & Few-shot NLU tasks with encoders \\
  & Meta-Learning (\autoref{par:meta-learn}) & Few labeled examples in related tasks & Adapts quickly to new tasks & Computationally intensive, potentially complex to implement & Few-shot classification tasks with encoders \\
  & Intermediate fine-tuning (\autoref{par:interm-ft}) & Data similar to final task & Improves convergence and final task performance & Sensitive to the similarity with the final task & Bridge PT and low-resource task (encoder)\\
  \multirow{-12}{*}{\rotatebox[origin=c]{90}{{Domain \& Cross-lingual FT \autoref{sec:ft-generalization}}}} & Multi-task learning (\autoref{sec:multi-task}) & Large, diverse set of related task data & Strong generalization to unseen tasks & High computational and data requirements & General-purpose generative models for zero-shot tasks\\
  \midrule
  & Embedding learning (\autoref{sec:embed-learn}) & Parallel or non-parallel data & Better transferability and semantic relationships & Static embeddings lack context & Discrepant pre-trained and target domain or language \\ 
  & Contrastive learning (\autoref{sec:contrastive}) & Paired data & Useful representations, data-efficiency & Sensitive to data pair quality and quantity  & Cross-lingual alignment; Classification tasks \\ 
  & Adversarial learning (\autoref{sec:adversarial}) & Unpaired data & Learns generalized features & Adversarial training is prone to be unstable & Cross-domain and cross-lingual transfer   \\ 
  & Semi-supervised learning (\autoref{sec:semi-supervised}) & Labeled \& unlabeled data & Utilizes unlabeled data, cost-effective & Sensitive to pseudo-label quality or noisy data & Classification and seq2seq low-resource tasks \\ 
  & Unsupervised learning (\autoref{sec:unsupervised}) & Unpaired data & No need for labeled data, scalable & Often worse performance than supervised & Scenarios with no labeled data \\ 
  \multirow{-12}{*}{\rotatebox[origin=c]{90}{{FT for Specialization \autoref{sec:ft-specialization}}}} & Active learning (\autoref{sec:active-learn}) & Labeled \& unlabeled data & Maximizes utility of labeled data & Requires iterative training and data selection & Classification and seq2seq low-resource tasks \\
  \bottomrule
  \end{tabular}
  \caption{Overview of the domain, cross-lingual, and specialization fine-tuning approaches discussed in this paper.}
  \label{tab:ft-methods}
\end{table*}

With the rapid growth of language models in recent years and the vast knowledge accumulated during large-scale pre-training (PT), research has increasingly focused on how to best shape and adapt this knowledge through FT for improved generalization across diverse tasks both for encoder \citep{schick_exploiting_2021,schick_its_2021,vu-etal-2020-exploring} and decoder models \citep{ouyang_training_2022,brown_language_2020}. 
For a concise summary of the methods, see \autoref{tab:ft-methods}.

\subsection{Continued pre-training} \label{sec:continued-pretrain}
Continued pre-training (CPT) bridges the gap between general PT and specialized domains or languages using unlabeled data. It reuses the original self-supervised PT objective \citep[or combines different ones for added capabilities like fill-in-the-middle;][]{bavarian_efficient_2022} allowing adaptation without supervised labels \citep{gururangan_dont_2020}, facilitating vocabulary extension and sample-efficient reuse of labeled FT data \citep{gnehm_evaluation_2022}.

For \textbf{cross-lingual} settings, aligning over diverse language data helps improve low-resource performance without considerably degrading high-resource capabilities \citep{imani_glot500_2023,blevins_language_2022}. 
Parallel corpora allow further gains via translation-based or other cross-lingual masking strategies \citep{wang_towards_2023,conneau_cross-lingual_2019}.
However, as the number of languages increases, the curse of dimensionality can dilute representational capacity, making language-specific adaptation or modular training strategies increasingly important \citep{alabi_adapting_2022,conneau_unsupervised_2020}.
For \textbf{domain adaptation}, CPT on modest amounts of high-quality in-domain text enables models to better capture domain-specific terminology and discourse patterns \citep{gururangan_dont_2020,bai_pre-train_2021}.
This approach typically outperforms supervised FT alone and is more effective than mixing domain-specific with general data and training from scratch \citep{turkmen_bioberturk_2023}.
Targeting specific vocabulary in the replaced or next token prediction objectives can further enhance performance \citep{lu_prompt_2023}.
In low-resource scenarios, proxy corpora or mixed-domain CPT can help, but the relevance of the data often outweighs its quantity \citep{mahapatra_entity_2022,jantscher_information_2023}.
In contrast to full encoder PT on billions or trillions of tokens, CPT with as few as 100K tokens can be beneficial, particularly for the newly introduced weights in PEFT methods \citep{jukic_parameter-efficient_2023,gnehm_evaluation_2022}.

Catastrophic forgetting remains a risk, especially for longer training runs with decoders \citep{kalajdzievski_scaling_2024}, requiring regularization techniques (\autoref{par:catastrophic-forget}) combined with PEFT methods (\autoref{sec:param-eff}) to preserve prior knowledge while adapting selectively \citep{liu_few-shot_2022,jukic_parameter-efficient_2023}.
In contrast, encoder models benefit even from longer training on small in-domain corpora \citep{gnehm_evaluation_2022}.
Scaling laws for decoder model PT suggest equal scaling of PEFT parameters and data size \citep{kaplan_scaling_2020, hoffmann_training_2022}.
Crucially, careful attention to (inconspicuous) data contamination is essential, especially for public datasets, as the overlap between PT and evaluation sets can artificially inflate results \citep{ruder_xtreme-up_2023}.

\subsection{Fine-tuning encoders and decoders}
While encoder and decoder architectures differ in their PT objectives (masked vs. autoregressive language modeling), the challenge of adapting them with limited data shares many similarities. 
Both require strategies that enhance generalization, mitigate overfitting, and leverage inductive biases from PT. 
However, some methods naturally align more with one model type: encoder-oriented techniques often operate in the representation space, while decoder-oriented ones emphasize task diversity and instruction-following. 
This section surveys approaches for both, highlighting where they converge and where their applicability diverges.

\paragraph{Pattern-exploiting training} \label{par:pattern-exploit}
(PET) reformulates classification as cloze-style language modeling for encoder models, aligning PT and FT objectives \citep{schick_exploiting_2021, schick_its_2021}. 
It relies on handcrafted patterns and verbalizers to map inputs and labels to token sequences, making it effective for few-shot tasks in low-resource domains and languages \citep{lu_prompt_2023,ullah_comparing_2023,song_taxonprompt_2023,qi_enhancing_2022}. 
Despite its success, inference often involves costly autoregressive decoding. 
Recent extensions mitigate this by introducing prototypical decoding \citep{karimi_mahabadi_prompt-free_2022}, label embedding learning, and evolutionary verbalizer optimization \citep{ling_evolutionary_2023}.
While PET originated in the encoder setting, its principles (of using prompts to bridge PT and downstream objectives) have influenced instruction-tuning methods for decoders.

\paragraph{Meta-learning.} \label{par:meta-learn}
Another line of encoder-focused research improves generalization by learning inductive biases that facilitate rapid adaptation to new discriminative tasks. 
Metric-based methods encode class-specific prototypes in an embedding space for non-parametric inference \citep{snell_prototypical_2017, wen_enhanced_2021}, while contrastive and triplet-based strategies further refine the latent space \citep{wu_improving_2024, pauli_anchoring_2023}. 
Optimization-based methods, such as model-agnostic meta-learning, learn initialization parameters in a two-level optimization that can quickly adapt to new tasks with few updates \citep{finn_model-agnostic_2017, bansal_self-supervised_2020, li_few-shot_2022, huang_meta-prompt_2023, chien_learning_2023}. 
Although less common today, meta-learning’s emphasis on representation regularization and rapid adaptation continues to influence newer methods that operate in the representation space of both encoder and decoder models.

\paragraph{Intermediate fine-tuning.} \label{par:interm-ft}
Encoders also benefit from intermediate FT (or task-adaptive PT), where models are fine-tuned on related, label-rich tasks before the target task.
This improves transfer by injecting domain-relevant inductive signals \citep{pruksachatkun-etal-2020-intermediate, vu-etal-2020-exploring,phang_sentence_2019}.
A closely related idea in decoder models is multi-task instruction tuning, where the model is exposed to a wide variety of datasets and instructions, scaling the benefits of intermediate supervision to heterogeneous task distributions. 
This distinction reflects the architectures: encoders profit from staged specialization, whereas decoders gain generalization by broad coverage of instructions.


\paragraph{Multi-task instruction tuning.} \label{sec:multi-task}
For decoder models, multi-task instruction tuning is the dominant paradigm for generalization.
By fine-tuning on diverse datasets and instruction styles, models acquire robust zero- and few-shot instruction-following abilities \citep{wei_finetuned_2021,aribandi_ext5_2021,sanh_multitask_2022}.
Compared to encoder-style intermediate fine-tuning, this approach is broader in scope, exposing decoders to heterogeneous task mixtures that yield checkpoints with faster convergence and stronger generalization to unseen tasks \citep{longpre_flan_2023,liu_few-shot_2022}. 
However, task selection and mixing require care, as diversity can improve generalization but also cause negative task transfer \citep{wang_how_2023,xia_less_2024}.







\paragraph{Mixture of experts.}
To mitigate such trade-offs, mixture-of-experts (MoE) architectures have emerged as a modular alternative for decoders.
Inputs are routed to specialized subnetworks based on input semantics \citep{shazeer_outrageously_2017,fedus_switch_2022}, via retrieval \citep{jang_exploring_2023} or dynamic gating mechanisms \citep{ostapenko_towards_2024,ponti_combining_2022}, improving modularity and continual learning. 
MoE systems often outperform monolithic instruction-tuned baselines in zero- and few-shot generalization, especially for compositional reasoning \citep{muqeeth_learning_2024}.
Recent large-scale decoders combine instruction tuning with massive MoE setups to reduce negative transfer while improving efficiency with fewer active parameters \citep{openai_gpt-oss-120b_2025,meta_llama4_2025}.

\paragraph{Effective instruction design.} \label{par:instruction-generalization}
Decoder performance is also sensitive to instruction formatting.
Generalization improves with prompt diversity, sequence packing, and careful task mixture balancing \citep{sanh_multitask_2022, iyer_opt-iml_2023}. 
Mixed prompting setups (zero-shot, few-shot, chain-of-thought) further enhance robustness \citep{longpre_flan_2023}. 
Inverting input-output roles and increasing lexical diversity acquire task-agnostic learning abilities \citep{min_metaicl_2022}.
These strategies not only equip models with robust in-context learning and reasoning abilities \citep[critical for low-shot generalization;][]{wei_chain--thought_2022, liu_pre-train_2021}, but also accelerate convergence in subsequent specialized FT \citep{longpre_flan_2023}.

\paragraph{To mask or not to mask?}
A subtle design choice in decoder fine-tuning is whether to apply loss only on output tokens or also on prompt tokens.
While the dominant practice masks loss on instruction tokens \citep{grattafiori_llama_2024,ouyang_training_2022,wei_finetuned_2021}, recent findings show that including prompt tokens as regularization improves generalization when instructions are long and data is limited \citep[<15k examples;][]{shi_instruction_2024,huerta-enochian_instruction_2024}.
Other forms of regularization, such as KL constraints against a base model may hurt instruction-following, while noise injection shows mixed results \citep{jain2024neftune}.

\section{Fine-tuning for Specialization} \label{sec:ft-specialization}
Fine-tuning for specialization focuses on adapting models to domain- or language-specific tasks that fall outside general capabilities (\autoref{tab:ft-methods}). 

\subsection{Embedding learning} \label{sec:embed-learn}
Since fixed vocabularies limit generalization to new domains and languages, recent work retrains or augments token embeddings (often with frozen transformer body) to better match target data distributions \citep{hung_tada_2023,artetxe_cross-lingual_2020}. 
This can bridge vocabulary gaps efficiently, particularly in low-resource or cross-lingual settings.
Subword-level tokenization remains standard for balancing efficiency and expressiveness \citep{kudo_sentencepiece_2018}, but it can fragment rare or domain-specific words. 
Entropy-based methods identify such tokens and selectively expand the vocabulary with new embeddings \citep{nag_entropy-guided_2023}. 
Alternative strategies map embeddings from high-resource to low-resource languages via seed dictionaries or aligned corpora \citep{mikolov_exploiting_2013,minixhofer_wechsel_2022}, or learn shared embedding spaces through alignment objectives \citep{cao_multilingual_2019,saadi_comparative_2022}.

\subsection{Contrastive and adversarial learning} 
Contrastive and adversarial objectives enhance adaptation by improving representation alignment across domains and languages and are compatible with PEFT methods (\autoref{sec:param-eff}).

\paragraph{Contrastive learning} \label{sec:contrastive}
pulls semantically similar pairs closer while pushing apart unrelated ones, improving both sentence- and word-level alignment in cross-lingual transfer \citep{chi_infoxlm_2021,chi_improving_2021,gaschi_exploring_2023,hu_language_2023}. It can be used on parallel corpora or via data augmentation for consistency, and integrates well with PEFT methods like adapters \citep{liu_grancats_2023,liu_intemats_2023,ahmat_wad-x_2023}. In downstream tasks, contrastive objectives boost performance for classification and similarity tasks using class descriptions, anchor prompts, or QA reformulations \citep{gao_simcse_2021,pauli_anchoring_2023,chen_few-shot_2023}.

\paragraph{Adversarial learning} \label{sec:adversarial}
improves generalization by training a discriminator that the model must deceive, encouraging domain- or language-invariant representations \citep{du_adversarial_2020,grieshaber_low-resource_2020,huang_improving_2023}. It is especially effective in the absence of parallel data and complements PEFT via language-specific adapters or domain-aware prefix tuning \citep{ngo_trung_unsupervised_2021,zhao_adpl_2022}.

\subsection{Limited supervision} \label{sec:limited-supervision}
In low-resource scenarios, semi-supervised, unsupervised, and active learning methods can successfully leverage unlabeled data to boost model generalization and robustness.

\paragraph{Semi-supervised learning} \label{sec:semi-supervised}
uses unlabeled data to augment small labeled sets. 
Common approaches include self-training with pseudo-labels filtered by model confidence or entropy \citep{schick_exploiting_2021,wang_boosting_2023,chen_mixtext_2020}, and consistency regularization to encourage stable predictions under perturbations \citep{sohn_fixmatch_2020,xie_unsupervised_2020}. 
Co-training variants train multiple views of the data to improve robustness \citep{clark_semi-supervised_2018}. 
These methods are particularly effective in low-resource adaptation, where pre-trained models provide valuable supervision signals to extend the limited annotated data.

\paragraph{Unsupervised methods} \label{sec:unsupervised}
often rely on consistency conditions such as input perturbation or cycle-consistency across modalities or domains \citep{zhu_unpaired_2017}. These methods enable learning bidirectional relationships without parallel data, useful for cross-domain, cross-lingual, or style transfer settings \citep{lample_phrase-based_2018,ren_unsupervised_2019,karisani_multiple-source_2022}. However, they typically require large unlabeled corpora and are most effective when combined with supervised signals.

\paragraph{Active learning} \label{sec:active-learn}
(AL) reduces annotation cost by selecting the most informative examples for labeling, typically using uncertainty (e.g., confidence, entropy, perplexity) or diversity (e.g., gradient similarity) criteria \citep{muradoglu_eeny_2022,yuan_cold-start_2020,francois_active_2023,karamcheti_mind_2021}. 
Iterative sampling and model re-initialization improve stability \citep{lemmens_combining_2023}.
AL reduces the amount of labeled data needed for classification task sometimes even to 30\% of the original size; can be beneficial above 1k samples, below any aditional labeled data works \citep{lemmens_combining_2023}.
It integrates well with PEFT methods like adapters and UniPELT, boosting sample efficiency in specialized tasks \citep{jukic_parameter-efficient_2023}.
The focus of AL on selective data efficiency aligns with findings that data quality often outweighs quantity in low-resource adaptation (see \autoref{par:influential-data} and \autoref{sec:pref_data}).

\subsection{Instruction tuning for specialization} \label{sec:instruction-tuning}
Instruction tuning specializes decoder models by aligning them with user intent and leveraging structured task descriptions.\footnote{We use \textbf{SFT} to denote supervised fine-tuning, with or without instructions.}
Beyond multi-task setups aimed at generalization (\autoref{sec:multi-task}), clear, specific, and contextually grounded instructions consistently improve task performance, even under severe data constraints \citep{wang_instructuie_2023,fleming_medalign_2024,zhang_multi-task_2023}. 
In contrast to the diverse \textbf{instruction design} required by multi-task models (\autoref{par:instruction-generalization}), instruction format variety only brings marginal gain in task-specific performance, and can even be counterproductive in some cases \citep{chen_maybe_2023}.
The lack of consistency between train and evaluation prompts can lead to performance drops, larger models being more robust to unseen formats \citep{gupta_instruction_2023}.

\paragraph{Specialization vs. generalization.}
Just as large-scale multi-task instruction tuning can hurt single-task performance, specializing too narrowly can reduce generalization \citep{wang_how_2023,zhang_when_2023,alabi_adapting_2022}. 
The best way to address this is by fine-tuning instruction-tuned models on the downstream task, which drastically improves sample efficiency, converges faster, and leads to overall better performance \citep{gupta_instruction_2023,longpre_flan_2023}. 
Adding general instruction data during SFT and applying regularization (\autoref{par:catastrophic-forget}) help maintain broader capabilities.





\paragraph{Scaling models and data.}
For multilingual closed-generation tasks (like translation or summarization), \citet{zhang_when_2023} find that SFT performance scales with data according to a power law with diminishing returns after an elbow point (around 10k–40k samples for PEFT), depending on the task.
SFT data has a larger impact on full FT than on PEFT, but full FT becomes justified after 100k to 1M samples. 
Surprisingly, model size has a larger impact than FT or PT data size and scaling PEFT parameters is virtually ineffective \citep{zhang_when_2023}.
The rationalization is that PEFT relies heavily on pre-trained knowledge, making model capacity more critical, especially for in-distribution tasks.
For more out-of-distribution or highly specific tasks, however, scaling fine-tuning data becomes more important, particularly in the early training phase.

\paragraph{Identifying influential data} \label{par:influential-data}
is key to developing specialized capabilities efficiently. 
Targeted instruction tuning can match full-data performance with as few as 10–15k samples, highlighting the value of high-quality supervision \citep{xia_less_2024,chen_maybe_2023}.
These approaches rely on statistical properties, similarity metrics, or gradient-based signals \citep{xie_data_2023,zhang_unreasonable_2018,xia_less_2024}, and show that a core subset remains effective across tasks and model scales. 
Beyond SFT, similar principles have been applied to filter CPT data based on task relevance \citep{xie_data_2023,mahapatra_entity_2022}, offering practical tools for denoising, data selection, or active augmentation (\autoref{sec:active-learn}).



\paragraph{Catastrophic forgetting mitigation.} \label{par:catastrophic-forget}
To prevent overfitting and preserve pre-trained knowledge during SFT with limited data, regularization techniques are vital. 
Early methods like layer-wise learning rate decay (LLRD), weight decay, and Mixout remain widely used, especially in encoder models \citep{howard_universal_2018,wiese_neural_2017,lee_mixout_2019}. 
For instruction tuning tasks with short outputs (e.g., classification, QA), applying a small loss weight (0.01–0.1) to the prompt tokens helps to avoid overfitting \citep{huerta-enochian_instruction_2024}. 
KL divergence is commonly used in LLM alignment to constrain outputs, but less so in standard SFT \citep{ouyang_training_2022}. 
Regardless of model size, learning rate warmup and small batch sizes consistently improve training stability with limited data \citep{buonocore_localizing_2023,xu_optimizing_2021}. 
Dropout and gradient clipping are also standard practices \citep{wolf-etal-2020-transformers,vonwerra2022trl}.


\paragraph{Hyperparameters for low-resource SFT}
Despite task and model differences, several practical defaults consistently yield strong performance in low-resource SFT. Learning rates between 5e-6 and 5e-5 are typical, with 2e-5 proving effective across popular models like Alpaca and Llama \citep{pareja_unveiling_2024,jin_rethinking_2023,grattafiori_llama_2024}. When tuning, scaling the learning rate by a factor of 3 is a simple yet effective heuristic.
Small batch sizes (1–8 per GPU) with gradient accumulation are favored for memory efficiency and better generalization \citep{ding_parameter-efficient_2023,xu_optimizing_2021}. 
While 2–3 epochs often suffice, training longer (up to 20-25 epochs) can help in extreme low-data regimes, provided early stopping is used to prevent overfitting \citep{muennighoff_scaling_2023,liu_natural_2024}.
Other stable choices include AdamW, cosine decay learning rate schedule with warmup ratios of 3–5\%, and gradient clipping set to 1. LoRA fine-tuning commonly uses rank and alpha values of 16 with 0.05 dropout \citep{vonwerra2022trl}.

\section{Preference Alignment} \label{sec:pref-align}
While fine-tuning equips PLMs with task- and domain-specific knowledge, it often falls short of aligning outputs with nuanced human preferences, especially when data is limited \citep{ganguli_red_2022,bai_training_2022}. 
Preference alignment (PA; \autoref{tab:pref-align}) addresses this gap by training models with ranked or scored feedback to better capture task-specific objectives like structure, helpfulness, truthfulness, or biases \citep{ziegler_fine-tuning_2020,christiano_deep_2017,ramamurthy_is_2022}.
This step usually follows supervised fine-tuning (SFT), once the model has already adapted to the task and domain. 
Without this prior adaptation, models tend to struggle with preference learning and exhibit higher rates of hallucination \citep{tunstall_zephyr_2024,ethayarajh_kto_2024,hong_orpo_2024}.
Refer to \autoref{tab:pref-align_methods} for a concise overview of preference alignment objectives. 

\begin{table*}[!ht]
  \centering
  \footnotesize
  \rowcolors{2}{white}{gray!10}
  \begin{tabular}{>{\centering}m{2.2cm}m{3.5cm}m{2.7cm}m{5.3cm}}
    \toprule
    \textbf{Method} & \textbf{Compute Characteristics} & \textbf{Data Requirements} & \textbf{Pros / Trade-offs} \\
    \midrule

    \nameref{par:pref_RL} 
    & High GPU demand; long training time; requires reward model and PPO
    & Ranked responses + reward model (100k+ samples)
    & Strong preference modeling, but sensitive to hyperparameters; expensive and unstable \\
    
    \nameref{par:pref_direct}
    & Moderate compute; fast convergence; no reward model; works with PEFT or full finetuning
    & Ranked or binary preference data (~10k+)
    & Simple pipeline; good performance; KTO uses weak binary signals; JPO compare instruction-response pairs; risk of verbosity exploitation \\
    
    \nameref{par:pref_ref-free}
    & Low compute; no reference model; efficient for PEFT; single-stage
    & Preference data + SFT data regularization (<10k)
    & Very efficient; no KL regularization needed; SimPO strong but lacks theory; vulnerable to reward hacking without regularization \\
    
    \nameref{par:pref_SFT}
    & Low GPU cost per step; single-stage (multiple iterations for SPIN)
    & Curated SFT data; optionally use rejection sampling
    & No preference data needed; limited by quality of SFT data; SPIN highly sample efficient \\
    
    \bottomrule
  \end{tabular}
  \caption{Taxonomy of preference alignment strategies with practical characteristics.}
  \label{tab:pref-align}
\end{table*}

  \begin{table*}[!ht]
  \centering
  \scriptsize
  \rowcolors{2}{gray!10}{white}
  \begin{tabular}{>{\cellcolor{white}}m{1mm}>{\centering\arraybackslash}m{1.1cm}>{\raggedright\arraybackslash}m{3.3cm}>{\raggedright\arraybackslash}m{7.4cm}>{\raggedright\arraybackslash}m{3.cm}}
  \toprule
  & \textbf{Method} & \textbf{Key idea} & \textbf{Training Objective} & \textbf{Hyperparameters} \\
  \midrule
  & RM+PPO \tiny\citep{ziegler_fine-tuning_2020} & 
  Learn a reward model and use it to guide output generation via RL & 
  $- r_\phi(x, y) + \beta \log \frac{\pi_\theta(y|x)}{\pi_{\text{ref}}(y|x)}$, where $r_\phi(x, y) = \mathbb{E}_{(x, y_w, y_l) \sim \mathcal{D}} \left[ \log \sigma \left( r_\phi(x, y_w) - r_\phi(x, y_l) \right) \right]$ & 
  $\beta \in \{0.001, 0.01, 0.05\}$ \\
  \multirow{-4}{*}{\rotatebox[origin=c]{90}{\footnotesize{\shortstack{RL \autoref{par:pref_RL}}}}} & d-RLAIF \tiny\citep{lee_rlaif_2024} & 
  Use LLM-generated preference scores as a reward to guide PPO training & 
  $- s(x, y) + \beta \log \frac{\pi_\theta(y|x)}{\pi_{\text{ref}}(y|x)}$, where $s(x, y)$ is the normalized score produced by an off-the-shelf LLM & 
  $\beta \in \{0.001, 0.01, 0.05\}$ \\
  \midrule
  & DPO \tiny\citep{rafailov_direct_2023} & 
  Prefer better responses via a logistic loss on reference-adjusted likelihoods & 
  $- \log \sigma \left( \beta \log \frac{\pi_\theta(y_w|x)}{\pi_{\text{ref}}(y_w|x)} - \beta \log \frac{\pi_\theta(y_l|x)}{\pi_{\text{ref}}(y_l|x)} \right)$ & 
  $\beta \in \{0.01, 0.025, 0.05, 0.1\}$\\
  & IPO \tiny\citep{azar_general_2024} & 
  Regularize DPO to avoid overfitting to extreme preference scores & 
  $\left( \log \frac{\pi_\theta(y_w|x)}{\pi_{\text{ref}}(y_w|x)} - \log \frac{\pi_\theta(y_l|x)}{\pi_{\text{ref}}(y_l|x)} - \frac{1}{2\tau} \right)^2$ & 
  $\tau \in \{0.01, 0.025, 0.0375, 0.5\}$ \\
  & R-DPO \tiny\citep{park_disentangling_2024} & 
  Penalize verbosity that can mislead preference learning & 
  $- \log \sigma \left( \beta \log \frac{\pi_\theta(y_w|x)}{\pi_{\text{ref}}(y_w|x)} - \beta \log \frac{\pi_\theta(y_l|x)}{\pi_{\text{ref}}(y_l|x)} + (\alpha |y_w| - \alpha |y_l|) \right)$ & 
  $\alpha \in \{0.05, 0.1, 0.5, 1.0\}$, $\beta \in \{0.01, 0.05, 0.1\}$ \\
  & KTO \tiny\citep{ethayarajh_kto_2024} & 
  Use unpaired binary feedback to optimize for coherence with reference distribution & 
  $- \lambda_w \sigma \left( \beta \log \frac{\pi_\theta(y_w|x)}{\pi_{\text{ref}}(y_w|x)} - z_{\text{ref}} \right) + \lambda_l \sigma \left( z_{\text{ref}} - \beta \log \frac{\pi_\theta(y_l|x)}{\pi_{\text{ref}}(y_l|x)} \right)$, where $z_{\text{ref}} = \mathbb{E}_{(x,y) \sim \mathcal{D}} [\beta \text{KL}(\pi_\theta(y|x) \| \pi_{\text{ref}}(y|x))]$ & 
  $\lambda_l = \lambda_w = 1.0$, $\beta \in \{0.01, 0.05, 0.1\}$ \\
  \multirow{-14}{*}{\rotatebox[origin=c]{90}{\footnotesize{\shortstack{Direct \autoref{par:pref_direct}}}}} & JPO \tiny\citep{bansal_comparing_2025} & 
  Generalize preference comparison to arbitrary input pairs & 
  $- \log \sigma \left( \beta \log \frac{\pi_\theta(y_w|x_w)}{\pi_{\text{ref}}(y_w|x_w)} - \beta \log \frac{\pi_\theta(y_l|x_l)}{\pi_{\text{ref}}(y_l|x_l)} \right)$ & 
  $\beta = 0.01$ \\
  \midrule
  & RRHF \tiny\citep{yuan_rrhf_2023} & 
  Simple regularised ranking-based alignment with length normalization & 
  $\max \left( 0, -\frac{1}{|y_w|} \log \pi_\theta(y_w|x) + \frac{1}{|y_l|} \log \pi_\theta(y_l|x) \right) - \lambda \log \pi_\theta(y_w|x)$ & 
  $\lambda \in \{0.1, 0.5, 1.0, 10.0\}$ \\
  & SLiC-HF \tiny\citep{zhao_slic-hf_2023} & 
  Calibrate sequence likelihood with a regularized, ranked loss & 
  $\max \left( 0, \delta - \log \pi_\theta(y_w|x) + \log \pi_\theta(y_l|x) \right) - \lambda \log \pi_\theta(y_w|x)$ & 
  $\delta = 1, \lambda \in \{0.1, 0.5, 1.0, 10.0\}$ \\
  & CPO \tiny\citep{xu_contrastive_2024} & 
  Removing the reference from the DPO loss leads to an efficient upper bound & 
  $- \log \sigma \left( \beta \log \pi_\theta(y_w|x) - \beta \log \pi_\theta(y_l|x) \right) - \log \pi_\theta(y_w|x)$ & 
  $\lambda = 1.0$, $\beta \in \{0.01, 0.05, 0.1\}$ \\
  & ORPO \tiny\citep{hong_orpo_2024} & 
  Discourage undesired behaviors during SFT with an odds-ratio loss. & 
  $- \log p_\theta(y_w|x) - \lambda \log \left( \frac{p_\theta(y_w|x)}{1 - p_\theta(y_w|x)} - \log \frac{p_\theta(y_l|x)}{1 - p_\theta(y_l|x)} \right)$,  where $p_\theta(y|x) = \exp \left( \frac{1}{|y|} \log \pi_\theta(y|x) \right)$ & 
  $\lambda \in \{0.1, 0.5, 1.0, 2.0\}$ \\
  \multirow{-12}{*}{\rotatebox[origin=c]{90}{\footnotesize{\shortstack{Reference-free \autoref{par:pref_ref-free}}}}} & SimPO \tiny\citep{meng_simpo_2024} & 
  Align reward function with the generation metric via length-aware, reference-free reward margin & 
  $- \log \sigma \left( \frac{\beta}{|y_w|} \log \pi_\theta(y_w|x) - \frac{\beta}{|y_l|} \log \pi_\theta(y_l|x) - \gamma \right)$ & 
  $\beta \in \{2.0, 2.5\}$, $\gamma \in \{0.3, 0.5, 1.0, 1.2, 1.4, 1.6\}$ \\
  \midrule
  
  & CoH \tiny\citep{liu_chain_2023} & 
  Teach preference via causal modeling over concatenated bad → good examples & 
  $- \log \pi_\theta\left( y_{\text{bad}},\ y_{\text{good}} \mid x \right)$, where input is $(x,\ y_{\text{bad}},\ y_{\text{good}})$ & 
  -- \\
  \multirow{-4}{*}{\rotatebox[origin=c]{90}{\footnotesize{\shortstack{SFT \autoref{par:pref_SFT}}}}} & SPIN \tiny\citep{chen_self-play_2024} & 
  Iteratively fine-tune using self-play to match human-like responses & 
  $- \log \sigma \left( \lambda \log \frac{\pi_\theta(y_w|x)}{\pi_{\theta_t}(y_w|x)} - \lambda \log \frac{\pi_\theta(y|x)}{\pi_{\theta_t}(y|x)} \right)$ & 
  $\lambda \in \{5.0 \text{ (last iter.)},\ 0.1 \text{ (else)}\}$ \\
  \bottomrule
  \end{tabular}
  \caption{Objectives and design motivations of various preference alignment strategies, with representative hyperparameters. See \autoref{sec:pref-align} dor detailed discussions.}
  \label{tab:pref-align_methods}
\end{table*}

\paragraph{Alignment with RL.} \label{par:pref_RL}
Reinforcement Learning from Human Feedback (RLHF) aligns model behavior by training a reward model to score outputs based on human preferences, followed by optimizing the model using RL algorithms like PPO \citep{ouyang_training_2022,ziegler_fine-tuning_2020,schulman_proximal_2017}. 
While effective, this process is resource-intensive (requiring large-scale human annotations, careful hyperparameter tuning, and substantial compute), making it inaccessible for most organizations \citep{touvron_llama_2023,rafailov_direct_2023,gao_scaling_2023}.
Variants like RLAIF replace human-labeled rewards with scores from an LLM during training, reducing cost while maintaining comparable performance \citep{lee_rlaif_2024}. However, RLAIF can be less coherent than RLHF, and both methods often produce longer outputs than SFT models, with occasional hallucinations.
Despite RLHF’s strengths, its instability and cost have spurred interest in more stable and efficient PA alternatives.

\paragraph{Direct preference alignment.} \label{par:pref_direct}
Recent methods simplify PA by eliminating the need for separate reward models and RL. 
Direct Preference Optimization \citep[DPO;][]{rafailov_direct_2023} reframes alignment as a binary classification problem over preference pairs, avoiding LM sampling, reward model training, and heavy hyperparameter tuning.
Several extensions build on DPO to improve generalization or efficiency. Identity Preference Optimization \citep[IPO;][]{azar_general_2024} introduces regularization to reduce overfitting. 
Kahneman-Tversky Optimization \citep[KTO;][]{ethayarajh_kto_2024} removes the need for paired preferences, relying only on binary feedback (good/bad). 
Length-regularized DPO \citep[R-DPO;][]{park_disentangling_2024} penalizes verbosity that may superficially align with human preferences. 
Joint Preference Optimization \citep[JPO;][]{bansal_comparing_2025} generalizes preference comparison beyond matching prompt inputs, capturing richer human judgments.
These approaches offer practical, scalable alternatives to RL-based methods while maintaining strong alignment performance.




\paragraph{Reference-free alignment} \label{par:pref_ref-free}
methods remove the need for a reference model by using alternative regularization strategies to keep outputs close to the SFT distribution and avoid degradation. 
This significantly reduces memory and compute costs, making these methods attractive for resource-constrained settings.
Contrastive preference optimization \citep[CPO;][]{xu_contrastive_2024} approximates DPO by replacing the reference model with a uniform prior, using standard NLL loss for regularization. 
ORPO \citep{hong_orpo_2024} merges SFT and alignment into a single step using an odds ratio loss. 
ORPO avoids the need for a reference model and separate FT step, offering greater efficiency than DPO or RLHF.
SimPO \citep{meng_simpo_2024} addresses misalignment between preference signals and generation behavior by using a normalized, reference-free reward based on likelihood and output length. A reward margin further sharpens the model’s ability to distinguish between preferred and rejected responses.
These methods offer lightweight alternatives to DPO and RLHF, especially suited for practical deployments in constrained environments.

\paragraph{Alignment during SFT.} \label{par:pref_SFT}
Another direction integrates PA directly into the SFT stage, removing the need for additional alignment steps. 
Careful curation of SFT datasets, such as filtering for specific structure or style, can yield strong alignment performance on its own, without dedicated preference data \citep{zhou_lima_2023,li_alpacaeval_2023,dong_raft_2023,yuan_rrhf_2023}.
Chain of Hindsight \citep{liu_chain_2023} presents both a good and bad response during training and teaches the model to prefer the good one at inference time, embedding preference signals without explicit reward modeling.
SPIN \citep{chen_self-play_2024} uses self-play to iteratively fine-tune a model to distinguish human-written and self-generated responses, gradually aligning its output distribution to human-like behavior.
These methods rely on high-quality SFT data and can serve as effective precursors to alignment techniques like DPO, emphasizing that data quality (and not quantity) is often the limiting factor (\autoref{sec:pref_data}), and that well-filtered SFT can be a lightweight, effective form of alignment.

\subsection{Choosing an alignment strategy} \label{sec:pref-choose}
PPO offers strong alignment performance but is resource-intensive and unstable, requiring large datasets and extensive hyperparameter tuning \citep{ouyang_training_2022,rafailov_direct_2023}. 
This makes it impractical in many cases.
Nevertheless, reward modeling itself has been shown to be relatively robust to moderate levels of label noise \citep[up to 30\%;][]{shen_towards_2024}.

DPO and its variants eliminate the reward model and are more stable.
IPO, especially, is less prone to reward hacking, achieving lower KL divergence under the same constraints \citep{rafailov_scaling_2024}.
DPO scales better than PPO to models of size up to 405B, but might require regualrization \citep[e.g., additional negative-log-likelihood (NLL) term with a coefficient of 0.2;][]{grattafiori_llama_2024}.
Contrary to reward modeling, DPO variants can be susceptible to label noise, as rankings are rarely flipped during alignment \citep{chen_preference_2024}. 
KTO is more sample efficient by learning from weaker binary signals and outperforms DPO in noisy or imbalanced datasets, especially in the 1B–30B parameter range \citep{ethayarajh_kto_2024}. 
JPO improves data efficiency by directly comparing instruction-response pairs, although a comparison might be ambiguous.
In general, masking out formatting tokens is recommended in objectives of contrasting nature to avoid conflicting signals \citep{grattafiori_llama_2024}.

Reference-free methods further reduce compute requirements by dispensing with reference models, and can be particularly effective in low-resource settings. 
For instance, CPO pushes translation performance beyond DPO using just 22k samples \citep{xu_contrastive_2024}. 
ORPO is on par with DPO, but produces more consistent and diverse outputs \citep{hong_orpo_2024}. 
SimPO delivers superior performance across multiple benchmarks, although the underlying reasons for its effectiveness lack theoretical justification \citep{meng_simpo_2024}. 
Despite their efficiency, all reference-free approaches are more susceptible to reward hacking, especially when fine-tuning the full model \citep{chen_preference_2024}.

Combining alignment objectives with SFT loss improves stability and performance for reasoning or domain-specific tasks \citep{xu_contrastive_2024,meng_simpo_2024}.
Integrating PA into SFT presents a lightweight alternative that requires only high-quality instruction data \citep{zhou_lima_2023}.
SPIN, in particular, achieves DPO-level performance with less data, however, it requires multiple training iterations, trading memory efficiency for time \citep{chen_self-play_2024}.

In practice, DPO and its variants remain strong defaults for general use. Reference-free methods are ideal for compute-constrained settings but demand careful regularization. 
Alignment during SFT is attractive when preference data is unavailable and curation quality is high. 
\autoref{tab:pref-align} summarizes trade-offs between alignment techniques, while \autoref{tab:pref-align_methods} contrasts their training objectives.

\subsection{Preference data criteria} \label{sec:pref_data}

\paragraph{Format and availability.}
Preference datasets typically consist of chosen/rejected pairs derived from instruction-response prompts \citep{ouyang_training_2022,rafailov_direct_2023,meng_simpo_2024}. 
Some approaches extend this to comparisons across prompts (JPO) or simple binary signals on the output (KTO), but the overall availability remains limited, with most datasets ranging from thousands to 1M samples and predominantly in English \citep{argilla_preference_2024,liu_glghawesome-llm-human-preference-datasets_2025}.

\paragraph{Synthetic feedback as a scalable solution.}
Collecting high-quality human preference data is resource-intensive. A widely adopted alternative is using synthetic feedback from strong LLMs guided by curated principles \citep{bai_constitutional_2022}. This enables broader scalability while maintaining quality when human oversight is present. Methods like SPIN go further by generating negative examples automatically via self-play, relying solely on an initial set of desirable outputs \citep{chen_self-play_2024}.
KTO also facilitates feedback collection by supporting imbalanced binary labels instead of strict pairwise comparisons \citep{ethayarajh_kto_2024}.

When feedback is derived from strong teacher models (e.g., GPT-4 or Claude), interestingly, SFT has been observed to rival preference-based alignment methods like DPO or RL \citep{sharma_critical_2024}.
This can be partly due to insufficient diversity or a weak student model not exceeding the quality of the synthetic data during exploration.
Diminishing returns beyond a few thousand samples suggest the importance of selection and coverage rather than raw volume.

\paragraph{Data quality over quantity.}
Multiple studies consistently show that a small number of high-quality, diverse examples outperforms larger, noisier datasets \citep{zhou_lima_2023,shen_towards_2024}. 
For instance, SPIN matches its original 50k-sample performance using just 1.8k well-curated examples, and ORPO remains effective at 141B scale with only 7k samples \citep{argilla_rlhf_2024}. 
Similar trends hold across alignment methods: as few as 1k–10k diverse, well-labeled examples can outperform tens or hundreds of thousands of lower-quality instances \citep{rafailov_direct_2023, zhou_lima_2023, shen_towards_2024}. 
For certain desired behaviors, like adhering to output structure specifications or coherent multi-turn conversations, a handful of high-quality examples can remarkably boost alignment \citep{zhou_lima_2023}.
These results underscore the value of filtering and targeted data collection over naive scaling.

Both human-labeled and synthetic preference datasets frequently contain noise, contradictions, or intransitive preferences \citep{ethayarajh_kto_2024}. These issues can undermine alignment objectives unless mitigated by clear annotation guidelines and consistent labeling practices. Additionally, prompting strategies like CoT can help improve the reliability of synthetic data, though techniques like few-shot prompting yield mixed results \citep{lee_rlaif_2024}.

\subsection{Scaling dynamics for model and data} \label{sec:pref_scaling}
PA performance generally follows a logarithmic scaling trend with respect to both model size and dataset size \citep{gao_scaling_2023}. 
For RLHF, increasing the size of the reward or policy model yields diminishing returns, and more data leads to better generalization but with progressively smaller gains. 
Notably, larger models do not overfit faster nor diverge more from the initial policy, contradicting earlier intuitions.
A minimum dataset size threshold is required to surpass near-random alignment performance. 
For example, at least 2k preference pairs are necessary for PPO to move beyond baseline loss, independent of the reward model size.
Moreover, data scaling is generally more impactful than reward model scaling: a 2-fold increase in data is better than a 4-fold increase in model size. For instance, a 680M model trained on 16k (32k) samples outperforms a 3B model with only 8k (16k) \citep[see Figure 10 in][]{gao_scaling_2023}.

In both RLHF and direct alignment methods, higher KL divergence budgets tend to degrade performance, often even before a full epoch is completed. Models frequently reach optimal performance at partial epochs (e.g., \textasciitilde 23k samples) with lower KL constraints \citep{rafailov_scaling_2024}. Smaller models, under the same constraints, achieve higher KL divergence and exhibit reward hacking earlier, while models around 7B parameters or more show significantly improved sample efficiency and stability.
With limited capacity or overly strict KL constraints, models tend to over-exploit spurious cues like verbosity. 
Length regularization alone does not fully prevent such behaviors \citep{rafailov_scaling_2024}.



Over-saturating SFT with synthetic completions can lead to marginal gains on alignment benchmarks, as observed with LLaMA-7B performance plateauing beyond 5k examples \citep{sharma_critical_2024}. This suggests that, beyond a point, SFT may be insufficient to further improve alignment, especially when data quality is limited, highlighting the need for more targeted alignment methods.

\subsection{Effect on downstream task performance} \label{sec:pref_downstream}
PA can have mixed effects on downstream task performance, depending on the alignment method, model initialization, and task type.
While alignment methods often improve truthfulness, commonsense knowledge and reasoning likely due to overlap with preference data, they may slightly reduce performance on knowledge-intensive tasks \citep{meng_simpo_2024}.
For instance, mathematical reasoning performance can significantly deteriorate for methods without a regularization SFT loss term, like CPO or ORPO.
\citet{tunstall_zephyr_2024} report that DPO alignment starts to hurt downstream performance after a single epoch.

Preference-aligned models often produce longer outputs than those trained with SFT alone, even when the objective includes length regularization \citep{park_disentangling_2024, meng_simpo_2024}.
However, verbosity shows weak correlation with KL divergence, implying that human preferences also depend on factors like repetition and tone, which may affect downstream performance.
Overall, the effects of alignment strategies on downstream outcomes remain underexplored, particularly in relation to model scale, output style, and non-English settings, highlighting a need for more systematic evaluation.

\section{Discussion} \label{sec:discussion}

\paragraph{Encode or decode?}
Choosing the appropriate model architecture is critical for optimizing performance across NLP tasks, particularly in resource-constrained or domain-specific settings \citep{alabi_massive_2020}. 
While large decoder-only models such as GPT-4 or LLaMA can be adapted to any NLP task via prompt-based reformulations \citep{liu_pre-train_2021}, they often require substantial computational resources for training and inference.
In contrast, encoder-only models like RoBERTa or DeBERTa \citep{liu_roberta_2019,he_deberta_2020} are generally more efficient for tasks centered on natural language understanding (NLU), such as classification, named entity recognition, and extractive QA, often outperforming decoder models with far fewer parameters \citep{benayas_comparative_2024,qorib_are_2024, garcia-diaz_leveraging_2023,schick_exploiting_2021}.
Their bidirectional attention and dense token supervision (i.e., predicting multiple masked tokens per input sequence) during training enhances sample efficiency and makes them especially effective in low-resource scenarios \citep{ding_parameter-efficient_2023, karimi_mahabadi_prompt-free_2022}.
For instance, RoBERTa (335M) achieves performance close to T5 (11B) despite having an order of magnitude fewer parameters \citep{ding_parameter-efficient_2023}.
While large decoder models dominate many mainstream applications, encoder models remain highly effective in domains requiring deep semantic understanding, limited supervision, and efficient inference. Their fixed-length embeddings and fast processing make them well-suited for tasks like retrieval, reranking, and domain-specific classification. 
Ultimately, architecture choice should be driven by task requirements, the trade-off between understanding and generation needs, and resource constraints.
\autoref{tab:nlp_tasks} summarizes effective encoder-based approaches for low-resource discriminative NLP tasks.





\paragraph{Cross-lingual transfer} \label{par:cross-ling-trans}
remains essential for deploying LLMs in multilingual settings, especially low-resource languages. 
Strong zero-shot generalization has been observed even when some languages were minimally present during PT \citep{armengol-estape_multilingual_2022}. 
Transfer improves with language similarity \citep[via word order or sub-word overlap;][]{deshpande_when_2022}, and can further be enhanced using PEFT methods like adapters or prefix-tuning \citep{wu_towards_2023, tu_efficiently_2024}. 
Cross-lingual prompting offers a training-free strategy for low-resource languages, with English prompts yielding more consistent results in few-shot settings \citep{lin_few-shot_2022, huang_not_2023}.
Prompting the model to think in English before responding in the target language further improves task performance via CoT reasoning \citep{qin_cross-lingual_2023}.
Translating training data generally outperforms direct transfer or translating test data in domain-specific tasks \citep{gaschi_exploring_2023}.
Finally, SFT using multilingual prompts (not just English) leads to better cross-lingual generalization \citep{muennighoff_crosslingual_2023}.

\paragraph{Model merging}
has emerged as a practical and effective strategy to enhance LLM capabilities without requiring additional training or ensembling overhead. By operating directly in weight space, merging enables the combination of models trained under different configurations, such as various data, hyperparameters, or training objectives (e.g., CPT, SFT, DPO/ORPO) into a single, stronger model \citep{lu_fine-tuning_2025, goddard_arcees_2024}. Techniques like Model Soup \citep{wortsman_model_2022} and SLERP \citep{shoemake_animating_1985} allow smooth interpolation between models, often leading to better downstream performance than any individual component. Notably, model merging contributed to top-ranked models on the Open LLM Leaderboard, being frequently adopted for the state of the art \citep{grattafiori_llama_2024}. It has also proven effective for low-resource languages, where merging models with complementary capabilities outperforms conventional CPT followed by SFT \citep{tao_unlocking_2024}. However, such benefits appear limited in smaller models (e.g., <2B parameters), suggesting scaling is crucial for emergent behaviors \citep{lu_fine-tuning_2025}.

\section{Conclusion}
Fine-tuning language models under data-scarce conditions remains a central challenge for practitioners and researchers aiming to build high-performing systems without the extensive resources typically required for large-scale training.
This survey presents a structured and pragmatic overview of the entire post-pretraining development pipeline, reviewing and organizing methods across parameter-efficient fine-tuning, domain and cross lingual generalization, specialization, and preference alignment. We highlight promising use cases, trade-offs, and best practices grounded in current empirical evidence.
Our findings underscore that larger models are consistently more sample-efficient. Further, data quality remains more critical than quantity, supporting the view that most capabilities are learned during pre-training, with fine-tuning and alignment primarily shaping model behavior.
We emphasize that despite the growing dominance of decoder-based models, encoder models remain competitive for discriminative tasks, and hybrid architectures provide promising avenues (e.g. RAG).
In preference alignment, the lack of correlation between alignment loss and task performance \citep{rafailov_scaling_2024} points to a deeper calibration issue that requires better metrics and theoretical grounding.
Model merging can be surprisingly effective, yet the nonlinear interactions underlying this behavior are poorly understood, motivating further research.
Finally, we stress the importance of transparent reporting of datasets, training protocols, and evaluation metrics for reproducibility and fair comparison, especially for non-English and domain-specific tasks, where current benchmarks often fall short for assessing real-world applicability.

\section*{Acknowledgments}
The authors would like to sincerely thank the action editor, Sebastian Ruder, and the anonymous reviewers for their insightful and constructive comments, which significantly helped improve the clarity and practical relevance of this work. This research was supported by the Nemetschek Innovation Foundation and the Bavarian Ministry of Science, Research and Art.

\bibliography{tacl2021}
\bibliographystyle{acl_natbib}

\appendix

\section{Task-specific training methods} \label{sec:task-specific}
This section serves as justification and further contextualization for the task-specific suggestions in \autoref{tab:nlp_tasks}.

\begin{table*}[!htb]
  \centering
  \scriptsize
  \rowcolors{2}{gray!10}{white}
  \begin{tabular}{>{\cellcolor{white}}c|>{\cellcolor{white}\centering\arraybackslash}m{2.4cm}|>{\centering\arraybackslash}m{1cm}|>{\raggedright\arraybackslash}m{3.8cm}|>{\raggedright\arraybackslash}m{4.2cm}}
      \toprule
      \textbf{Task} & \textbf{Continued Pre-training} & \textbf{Labeled Data*} & \centering\arraybackslash\textbf{Fine-tuning} & \centering\arraybackslash\textbf{Complementary Options} \\
      \midrule
      & & < 250  & PET + Adapters & Label embeddings; Similar task data \\
      & & < 1K & NLI format with Prefix-tuning / UniPELT & Fine-tune first on high-resource NLI data \\
      & & < 10K & LoRA; LLRD & Active or Semi-supervised learning \\
      \multirow{-5}{*}{\rotatebox[origin=c]{90}{{\shortstack{Text\\Classification}}}} & \multirow{-5}{=}{75K+ tokens for PEFT params, LR warm-up + ER} & 10K+ & Full fine-tuning &  Regularization (small batches, gradient clipping) \\
      \midrule 
      & & < 250 & PET; QA format with CL & Intermediate training, CL between query/entity \\
      & & < 5K & Adapters, Parameter masking  & Similar task datasets for knowledge distillation, Active learning \\
      \multirow{-5}{*}{\rotatebox[origin=c]{90}{{\shortstack{Sequence\\Labeling}}}} & \multirow{-5}{=}{25K+ tokens for PEFT params, LR warm-up + ER + Mixout} & 5K+ & Full fine-tuning & LLRD; Bounded gradients\\
      \midrule
      & & < 250 & PET + Adapters / Prompt-tuning & CL with label embeddings, Data augmentation \\
      & & < 1K & Adapters / UniPELT & Multilingual data with CL \\
      & & < 10K & PEFT (LoRA) & CL on NLI data \\
      \multirow{-5}{*}{\rotatebox[origin=c]{90}{{\shortstack{Text\\Comparison}}}} & \multirow{-5}{=}{75K+ tokens for PEFT params, consistency CL, LR warm-up} & 10K+ & Full fine-tuning & Bounded gradients \\
      \bottomrule
  \end{tabular}
  \caption{Suggested approaches for NLP task groups with limited data. All information in this table was compiled from reviewed papers (\autoref{sec:task-specific}).
  Experience replay (ER) stands for mixing in some general-domain data as regularization.
  *Labeled data for Sequence Labeling is provided in sentences; for other tasks, in examples.}
  \label{tab:nlp_tasks}
\end{table*}

\subsection{Text classification}
In ultra-low-resource settings (2–8 examples per class), contrastive learning (CL) methods like few-shot sentence transformers trained with triplet loss and anchoring on class descriptions are effective \citep{pauli_anchoring_2023}. Reformulating classification as NLI improves few-shot performance by enabling transfer from general-purpose NLI datasets \citep{laurer_less_2024}, and is especially effective with <1k examples. For similar data sizes, pattern-exploiting training (PET) consistently outperforms full FT across models like RoBERTa, XLM-R, and DistilBERT on NLI tasks, topic/sentiment classification \citep{schick_exploiting_2021,ullah_comparing_2023, ling_evolutionary_2023}, and can be improved by using label embeddings and adapters \citep{karimi_mahabadi_prompt-free_2022}.

For <1k examples, prefix tuning and UniPELT outperform full FT with BERT, especially when combined with CPT on unlabeled data (~75k tokens) \citep{jukic_parameter-efficient_2023}. Soft prompt tuning is also superior in few-shot settings (2–64 shots), particularly with domain-specific models like ClinicalBERT and SciBERT \citep{goswami_switchprompt_2023}. Domain adaptation (CPT) with token millions, including vocabulary injection and last-layer tuning, further boosts performance on document classification \citep{gnehm_evaluation_2022}. 
Prefix tuning works well up to 1k samples, while LoRA improves between 1k–10k, and full FT becomes competitive beyond that \citep{mao_unipelt_2022}. Regularization methods like layerwise learning rate decay (LLRD) and mixout help stabilize training with BERT on GLUE tasks under 10k examples \citep{zhang_revisiting_2020}. Active learning becomes impactful above 1k samples and can reduce labeling needs by up to 70\% \citep{lemmens_combining_2023, ein-dor_active_2020, yuan_cold-start_2020}.

For higher-resource setups (\textasciitilde10k examples), full FT performs best, especially when paired with gradient clipping and small batch sizes for stability in reading comprehension \citep{xu_optimizing_2021}. Finally, for tasks like NLI, even a subset of 16k examples (1.9M tokens) selected through clustering and coreset sampling is sufficient to train competitive decoder models like Galactica 1.3B \citep{chen_maybe_2023}.

\subsection{Sequence labeling}
In extreme low-resource setups (1–10 examples), pseudo-labeling enables 1-shot NER by using predicted spans as supervision \citep{peng_less_2023}. Reformulating NER as QA and applying CL between the query and target entity (positive pair) can further enhance few-shot learning (5–50 shots) in biomedical NER. Transferring knowledge from high-resource QA tasks boosts performance \citep{chen_few-shot_2023}. Similarly, for biomedical relation extraction (RE), intermediate training with multiple related datasets and knowledge distillation from an ensemble outperforms PET, especially when task similarity (e.g., vocabulary overlap) is high \citep{moscato_multi-task_2023}.
In multilingual RE, mT5 (220M) underperforms XLM-R base (125M) despite its size, especially in languages with less pretraining data. For low-resource scenarios (<32 examples per class), prompts and label names should be kept in English for optimal transfer \citep{chen_multilingual_2022}.
At 100–1k sentences, CPT domain adaptation is particularly effective. Using as few as 36k tokens improves BERT-based NER, and selecting the most relevant data subset yields better performance than using full corpora \citep{mahapatra_entity_2022}. In German clinical RE/NER, FT with just 512–1400 sentences is feasible when combined with active learning (perplexity-based sampling) and domain-specific models like German-MedBERT or R-BERT \citep{jantscher_information_2023}.
In the 1k–5k sentence range, PEFT strategies remain effective. CPT domain-adaptation offers limited gains, as fine-tuning the base model sometimes performs equally well \citep{crema_advancing_2023}. Additional techniques like LLRD, mixout, and learning rate warm-up improve generalization for NER, RE, and QA across tasks under 8k sentences \citep{buonocore_localizing_2023}. Adapter-based extensions with limited FT (\textasciitilde5k sentences) also prove competitive when pretrained on large domain-specific corpora \citep{tai_exbert_2020}.

At moderate scale (5k–10k), BERT-based models trained from scratch on large biomedical corpora (~12B tokens) perform comparably to those adapted via CPT on 1B tokens, offering flexibility in resource usage \citep{bai_pre-train_2021}. 

\subsection{Text comparison}
CL proves to be a powerful pretraining strategy for improving semantic representations in tasks like NLI or paraphrase detection. Models like DeBERTa large (435M) benefit significantly from CL pretraining when used for zero- or few-shot textual entailment \citep{kowsher_contrastive_2023}. Similarly, SimCSE and ConSERT show that both unsupervised CL and supervised NLI-based learning greatly enhance sentence embeddings without labeled task data \citep{gao_simcse_2021,yan_consert_2021}.
For ultra-low supervision (16–64 examples), PEFT techniques consistently outperform full FT. RoBERTa large (355M), using a cloze-style format with adapters and label embedding learning, surpasses FT, BitFit, PET, and prompt tuning with just 16 samples per label \citep{karimi_mahabadi_prompt-free_2022}. PET also performs well for multi-level implicit discourse relation recognition using RoBERTa \citep{zhao_infusing_2023}.
In multilingual and cross-lingual comparison tasks, PET combined with CL consistency loss shows strong results. On tasks like XNLI and paraphrase identification, models such as XLM-R base (279M), XLM-R large (561M), and InfoXLM large (561M) benefit from this hybrid approach, especially in 16–256-shot learning settings \citep{qi_enhancing_2022}.
A CL framework using 500 parallel sentence pairs, supplemented with up to 10k non-parallel instances, establishes strong information retrieval performance across languages \citep{hu_language_2023}.
For 1k–10k instances, lightweight tuning strategies remain efficient: adapter-based tuning works well below 1k examples, while LoRA is effective for 1k–10k, as shown on BART\textsubscript{large} across text comparison tasks \citep{mao_unipelt_2022}.

\end{document}